%% file: icml_paper.tex
\documentclass{article}

\usepackage{times}
\usepackage{graphicx} 
\usepackage{subcaption}

\usepackage{natbib}

\usepackage{algorithm}
\usepackage{algorithmic}

\usepackage{hyperref}

\usepackage[accepted]{icml2017}

\usepackage{amsmath}
\usepackage{amsthm}
\usepackage{amssymb}
\usepackage{booktabs}
\usepackage{color}
\usepackage{mathabx}
\usepackage{wrapfig}
\usepackage{enumitem}

\input{macros}

\icmltitlerunning{Coordinated Multi-Agent Imitation Learning}

\begin{document} 

\twocolumn[
\icmltitle{Coordinated Multi-Agent Imitation Learning}



\icmlsetsymbol{equal}{*}

\begin{icmlauthorlist}
\icmlauthor{Hoang M. Le}{caltech}
\icmlauthor{Yisong Yue}{caltech}
\icmlauthor{Peter Carr}{disney}
\icmlauthor{Patrick Lucey}{stats}
\end{icmlauthorlist}

\icmlaffiliation{caltech}{California Institute of Technology, Pasadena, CA}
\icmlaffiliation{disney}{Disney Research, Pittsburgh, PA}
\icmlaffiliation{stats}{STATS LLC, Chicago, IL}

\icmlcorrespondingauthor{Hoang M. Le}{hmle@caltech.edu}
\icmlkeywords{boring formatting information, machine learning, ICML}

\vskip 0.3in
]



\printAffiliationsAndNotice{}  

\begin{abstract} 
We study the problem of imitation learning from demonstrations of multiple coordinating agents. One key challenge in this setting is that learning a good model of coordination can be difficult, since coordination is often implicit in the demonstrations and must be inferred as a latent variable.  We propose a joint approach that simultaneously learns a latent coordination model along with the individual policies. In particular, our method integrates unsupervised structure learning with conventional imitation learning. We illustrate the power of our approach on a difficult problem of learning multiple policies for fine-grained behavior modeling in team sports, where different players occupy different roles in the coordinated team strategy.  We show that having a coordination model to infer the roles of players  yields substantially improved imitation loss compared to conventional baselines.
\end{abstract} 


\input{sec-intro}

\input{sec-problem}
\input{sec-method}


\input{sec-experiment}

\input{sec-related}

\input{sec-conclude}

\textbf{Acknowledgement.} This work was funded in part by NSF Awards \#1564330 \& \#1637598, JPL PDF IAMS100224, a Bloomberg Data Science Research Grant, and a gift from Northrop Grumman.

\newpage

\begin{small}
\bibliography{icml_paper}
\bibliographystyle{icml2017}
\end{small}

\newpage

\appendix

\input{sec-derivation}

\end{document}

%% file: macros.tex
\renewcommand{\algorithmicrequire}{\textbf{Input:}}
\renewcommand{\algorithmicensure}{\textbf{Output:}}

\newcommand{\pivec}{\vec{\pi}}
\newcommand{\svec}{\vec{s}}
\newcommand{\avec}{\vec{a}}

\usepackage{forloop}
\newcounter{ct}

\newcommand{\E}{\mathbb{E}}

\newcommand{\train}{\texttt{Train}}

\graphicspath{{figures/}}

\theoremstyle{plain}

\theoremstyle{definition}

\theoremstyle{remark}

%% file: sec-intro.tex
\section{Introduction}
\label{introduction}
The areas of multi-agent planning and control have witnessed a recent wave of strong interest due to the practical desire to deal with complex real-world problems, such as smart-grid control, autonomous vehicles planning, managing teams of robots for emergency response, among others. 
From the learning perspective, (cooperative) multi-agent learning is not a new area of research \cite{stone2000multiagent,panait2005cooperative}. However, compared to the progress in conventional supervised learning and single-agent reinforcement learning, the successes of multi-agent learning have remained relatively modest. Most notably, multi-agent learning suffers from extremely high dimensionality of both the state and actions spaces, as well as relative lack of data sources and experimental testbeds. 


\begin{figure}[tb]
	\centering
	\includegraphics[scale=0.45]{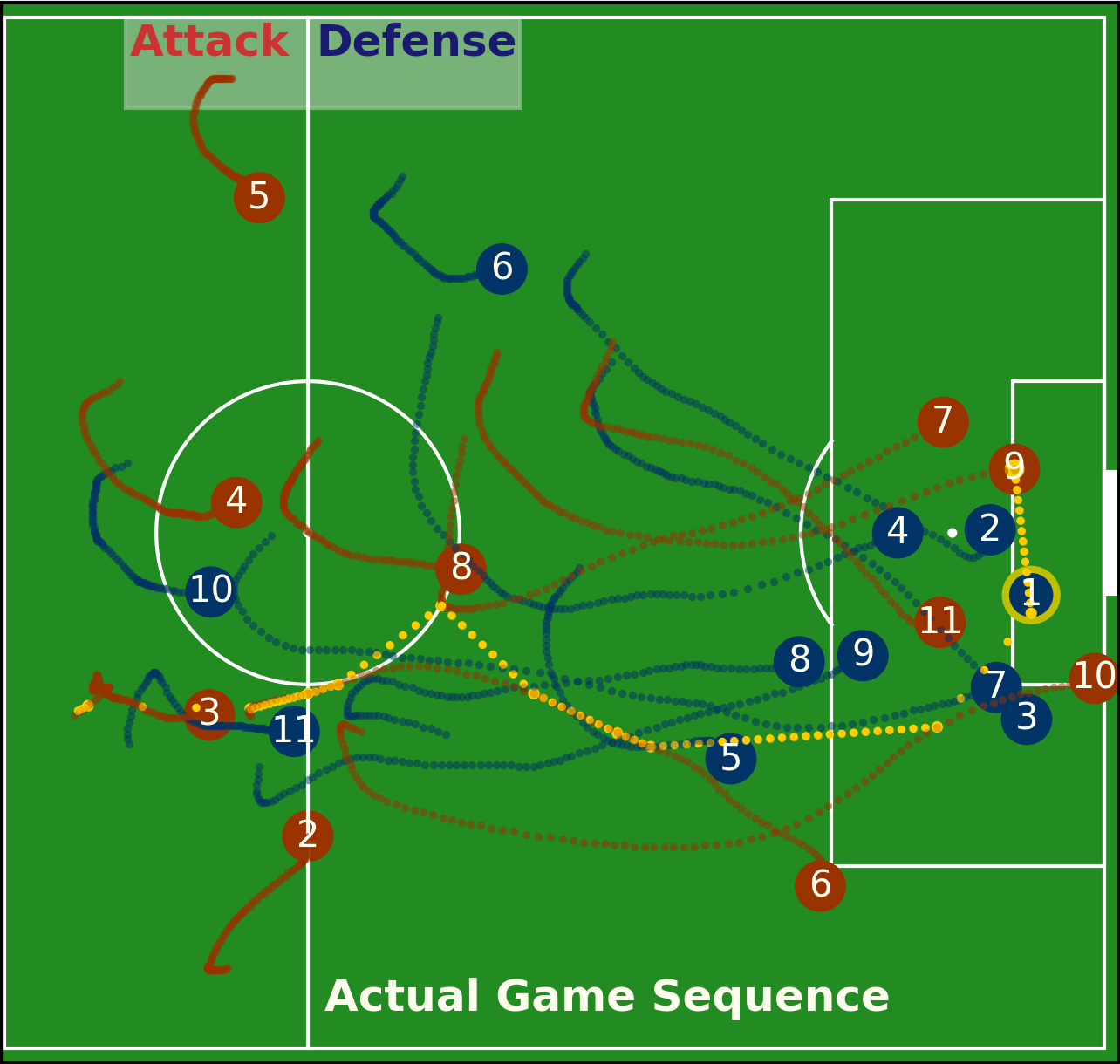}
	\vspace{-0.1in}
	\caption{\textit{Our motivating example of learning coordinating behavior policies for team sports from tracking data. Red is the attacking team, blue is the defending team, and yellow is the ball. }
	\label{fig:game_example}}
\end{figure}

The growing availability of data sources for coordinated multi-agent behavior, such as sports tracking data \cite{formation}, now enables the possibility of learning multi-agent policies from demonstrations, also known as multi-agent imitation learning.  One particularly interesting aspect of domains such as team sports is that the agents must coordinate.  For example, in the professional soccer setting depicted in Figure \ref{fig:game_example}, different players must coordinate to assume different roles (e.g., defend left field).  However, the roles and role assignment mechanism are unobserved from the demonstrations.  Furthermore, the role for a player may change during the same play sequence.  In the control community, this issue is known as ``index-free'' multi-agent control \cite{kingston2010index}.

Motivated by these challenges, we study the problem of imitation learning for multiple coordinating agents from demonstrations. Many realistic multi-agent settings require coordination among collaborative agents to achieve some common goal \cite{guestrin2002coordinated,kok2003multi}. Beyond team sports, other examples include learning policies for game AI, controlling teams of multiple robots, or modeling collective animal behavior. As discussed above, we are interested in settings where agents have access to the outcome of actions from other agents, but the coordination mechanism is neither clearly defined nor observed, which makes the full state only partially observable. 

We propose a semi-supervised learning framework that integrates and builds upon conventional imitation learning and unsupervised, or latent, structure learning. The latent structure model encodes a coordination mechanism, which approximates the implicit coordination in the demonstration data.
In order to make learning tractable, we develop an alternating optimization method that enables integrated and efficient training of both individual policies and the latent structure model. 
For learning individual policies, we extend reduction-based single-agent imitation learning approaches into multi-agent domain, utilizing powerful black-box supervised techniques such as deep learning as base routines. 
For latent structure learning,  we develop a stochastic variational inference approach. 

We demonstrate the effectiveness of our method in two settings.  The first is a synthetic experiment based on the popular predator-prey game. The second is a challenging task of learning multiple policies for team defense in professional soccer, using a large training set\footnote{Data at \url{http://www.stats.com/data-science/} and see video result at \url{http://hoangminhle.github.io}}  of play sequences illustrated by Figure \ref{fig:game_example}. We show that learning a good latent structure to encode implicit coordination yields significantly superior imitation performance compared to conventional baselines.
To the best of our knowledge, this is the first time an imitation learning approach has been applied to jointly learn cooperative multi-agent policies at large scale. 


%% file: sec-problem.tex
\section{Problem Formulation} 
\label{problem}
In coordinated multi-agent imitation learning, we have $K$ agents acting in coordination to achieve a common goal (or sequence of goals). Training data $\mathcal{D}$ consists of multiple demonstrations of $K$ agents.  Importantly, we assume the identity (or indexing) of the $K$ experts may change from one demonstration to another. Each (unstructured) set of demonstrations is denoted by $U=\{U_{1},\ldots, U_{K}\}$, where $U_k = \{ u_{t,k}\}_{t=1}^T$ is the sequence of actions by agent $k$ at time $t$.  Note that each set of demonstrations can have varying sequence length T. Let $C = \{ c_t\}_{t=1}^T$ be the context associated with each demonstration sequence. 

\textbf{Policy Learning.} Our ultimate goal is to learn a (largely) decentralized policy, but for clarity we first present the problem of learning a fully centralized multi-agent policy.  Following the notation of \cite{dagger}, let $\pivec(\svec) := \avec$ denote the joint policy that maps the joint state, $\svec = [s_1,\ldots,s_K]$, of all $K$ agents into $K$ actions $\avec = [a_1,\ldots,a_K]$.   The goal is to minimize imitation loss:
\vspace{-3pt}
$$\mathcal{L}_{imitation} = \E_{\svec \sim d_{\pivec}}\left[ \ell(\pivec(\svec)) \right],$$
where $d_{\pivec}$ denotes the distribution of states experienced by joint policy $\pivec$ and $\ell$ is the imitation loss defined over the demonstrations (e.g., squared loss for deterministic policies, or cross entropy for stochastic policies).    

The decentralized setting decomposes the joint policy $\pivec = [\pi_1,\ldots,\pi_K]$ into $K$ policies, each tailored to a specific agent index or ``role''.\footnote{It is straightforward to extend our formulation to settings where multiple agents can occupy the same role, and where not all roles are occupied across all execution sequences.}  The loss function is then:
\vspace{-3pt}
$$\mathcal{L}_{imitation} = \sum_{k=1}^K\E_{s \sim d_{\pi_k}}\left[ \ell(\pi_k(s_k)) \right].$$

\textbf{Black-Box Policy Classes.} In order to leverage powerful black-box policy classes such as random forests and deep learning, we take a learning reduction approach to training $\pivec$.  One consequence is that the state space representation  $s=[ s_1, \ldots, s_K]$ must be consistently indexed, e.g., agent $k$ in one instance must correspond to agent $k$ in another instance.  This requirement applies for both centralized and decentralized policy learning, and is often implicitly assumed in prior work on multi-agent learning. A highly related issue arises in distributed control of index-free coordinating robots, e.g., to maintain a defined formation \cite{kloder2006path,kingston2010index}.

\textit{\textbf{Motivating example: Soccer Domain.}} Consider the task of imitating professional soccer players, where training data includes play sequences from different teams and games. Context $C$ corresponds to the behavior of the opposing team and the ball. The data includes multiple sequences of $K$-set of trajectories $U = \{ U_1,U_2,\ldots, U_K\}$, where the actual identity of player generating $U_k$ may change from one demonstration to the next.

One important challenge for black-box policy learning is constructing an indexing mechanism over the agents to yield a consistent state representation. For example, the same index should correspond to the ``left defender'' in all instances. Otherwise, the inputs to the policy will be inconsistent, making learning difficult if not impossible. Note that barring extensive annotations or some heuristic rule-based definitions, it is unnatural to quantitatively define what makes a player ``left defender''. In addition, even if we had a way to define who the ``left defender'' is, he may not stay in the same role during the same sequence. 

\textbf{Role-based Indexing.} We address index-free policy learning via role learning and role-based index assignment. To motivate our notion of role, let's first consider the simplest indexing mechanism: one could equate role to agent identity. However, the data often comes from various sequences, with heterogeneous identities and teams of agents. Thus instead of learning identity-specific policies, it is more natural and data-efficient to learn a policy per role. However, a key challenge in learning policies directly is that \textit{the roles are undefined, unobserved, and could change dynamically within the same sequence.} We thus view learning the coordination, via role assignment, as an unsupervised structured prediction problem.


\textbf{Coordination via Structured Role Assignment.}
Instead of handcrafting the definition of roles, we learn the roles in an unsupervised fashion, without attaching any semantic labels to the roles. At the same time, role transition should obey certain structural regularity, due to coordination. This motivates using graphical models to represent the coordination structure. 

\textbf{Coordinated Policy Learning.} We formulate the indexing mechanism as an assignment function $\mathcal{A}$ which maps the unstructured set $U$ and some probabilistic structured model $q$ to an indexed set of trajectory $A$ rearranged from $U$, i.e.,
\begin{equation*}
\mathcal{A}:\{ U_1,..,U_K\}\times q\mapsto \left[ A_1,..,A_K\right],
\end{equation*}
where the set $\{ A_1,..,A_K\} \equiv \{ U_1,..,U_K\}$.
We view $q$ as a latent variable model that infers the role assignments for each set of demonstrations. Thus, $q$ drives the indexing mechanism $\mathcal{A}$ so that state vectors can be  consistently constructed to facilitate optimizing for the imitation loss.

We employ entropy regularization, augmenting the imitation loss with some low entropy penalty \cite{grandvalet2004semi,dudik2004performance}, 
   yielding our overall objective:
\begin{equation}
\label{eqn:objective}
\min_{\pi_1,..,\pi_K, \mathcal{A}} \sum_{k=1}^K \E_{s_k \sim d_{\pi_k}} \left[ \ell(\pi_k(s_k)) | \mathcal{A}, \mathcal{D} \right] - \lambda H(\mathcal{A}|\mathcal{D})
\end{equation}
where both imitation loss and entropy are measured with respect to the state distribution induced by the policies, and $\mathcal{D}$ is training data. This objective can also be seen as maximizing the mutual information between latent structure and observed trajectories \cite{krause2010discriminative}.

%% file: sec-method.tex
\begin{figure}[t]
	\centering
	\includegraphics[scale=0.38]{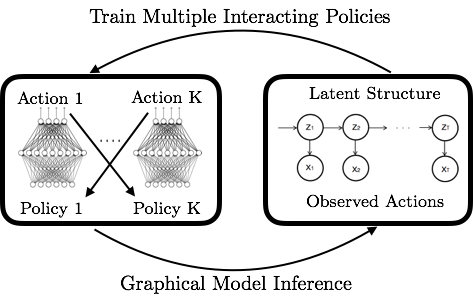}
	\vspace{-0.1in}
	\caption{\textit{Alternating stochastic optimization training scheme for our semi-supervised structure regularization model.}} 
	\label{fig:scheme}
\end{figure}

\section{Learning Approach}
\label{method}
Optimizing \eqref{eqn:objective} is challenging for two reasons. First, beyond the challenges inherited from single-agent settings, multi-agent imitation learning must account for multiple simultaneously learning agents, which is known to cause non-stationarity for multi-agent reinforcement learning \cite{busoniu2008comprehensive}. Second, the latent role assignment model, which forms the basis for coordination, depends on the actions of the learning policies, which in turn depend on the structured role assignment. 

We propose an alternating optimization approach to solving \eqref{eqn:objective}, summarized in Figure \ref{fig:scheme}. The main idea is to integrate imitation learning with unsupervised structure learning by taking turns to (i) optimize for imitation policies while fixing a structured model (minimizing imitation loss), and (ii) re-train the latent structure model and reassign roles while fixing the learning policies (maximizing role assignment entropy). The alternating nature allows us to circumvent the circular dependency between policy learning and latent structure learning. Furthermore, for (i) we develop a stable multi-agent learning reduction approach.

\subsection{Approach Outline} 
\label{subsec:combined_method}
Algorithm \ref{algo:main_algo} outlines our framework. We assume the latent structure  model for computing role assignments is formulated as a graphical model. The multi-agent policy training procedure $\texttt{Learn}$ utilizes a reduction approach, and can leverage powerful off-the-shelf supervised learning tools such as deep neural networks \cite{hochreiter1997long}. The structure learning $\texttt{LearnStructure}$ and role assignment $\texttt{Assign}$ components are based on graphical model training and inference. For efficient training, we employ alternating stochastic optimization \cite{hoffman2013stochastic,johnson2014stochastic,beal2003variational} on the same mini-batches. Note that batch training can be deployed similarly, as illustrated by one of our experiments. 


\begin{algorithm}[tb]
	\begin{small}
	\caption{ Coordinated Multi-Agent Imitation Learning} 
	\label{algo:main_algo}
	\begin{algorithmic}[1]
		\REQUIRE Multiple unstructured trajectory sets $U = \{ U_1,\ldots, U_K\}$ with $U_k = \{u_{t,k}\}_{t=1}^T$ and context $C = \{ c_t\}_{t=1}^T$. 
        \REQUIRE Graphical model $q$ with global/local parameters $\theta$ and $z$.
        \REQUIRE Initialized policies $\pi_k, k=1,\ldots,K$
        \REQUIRE Step size sequence $\rho_n, n=1,2,\ldots$
		\REPEAT
		\STATE $[A_1,\ldots,A_K] \leftarrow \texttt{Assign}\{U_1,\ldots,U_K \vert q(\theta,z)\}$ \\
		\STATE $[\pi_1,\ldots,\pi_K] \leftarrow \texttt{Learn}\left[A_1,\ldots,A_K,C \right]$
		\STATE Roll-out $\pi_1,\ldots, \pi_K$ to obtain $\widehat{A}_1,\ldots,\widehat{A}_K$
		\STATE $A_k \leftarrow \widehat{A}_k \enskip \forall k$ \\
		(Alternatively: $A_k \leftarrow \widehat{A}_k$ with prob $\eta$ for $\eta\rightarrow 1$)
		\STATE $q(\theta,z) \leftarrow \texttt{LearnStructure}\{ A_1,\ldots,A_K,C,\theta,\rho_n\} $
		\UNTIL No improvement on validation set
		\OUTPUT $K$ policies $\pi_1,\pi_2,\ldots,\pi_K$
	\end{algorithmic}
	\end{small}
\end{algorithm}

We interleave the three components described above into a complete learning algorithm. Given an initially unstructured set of training data, an initialized set of policies, and prior parameters of the structure model, Algorithm \ref{algo:main_algo} performs alternating structure optimization on each mini-batch (size $1$ in Algorithm \ref{algo:main_algo}). 
\vspace{-0.1in}
\begin{itemize}
\item Line 2: Role assignment is performed on trajectories $\{A_1,\ldots,A_K\}$ by running inference procedure (Algorithm \ref{algo:assignment}). The result is an ordered set $\left[ A_1,\ldots, A_K\right]$, where trajectory $A_k$ corresponds to policy $\pi_k$.
\vspace{-0.05in}
\item Line 3-5: Each policy $\pi_k$ is updated using joint multi-agent training on the ordered set $\left[A_1,\ldots,A_K,C\right]$ (Algorithm \ref{algo:joint_imitation}). The updated models are executed to yield a rolled-out set of trajectories, which replace the previous set of trajectories $\{ A_k\}$.
\vspace{-0.05in}
\item Line 6: Parameters of latent structured model are updated from the rolled-out trajectories (Algorithm \ref{algo:structure_learning}).
\end{itemize}
\vspace{-0.1in}

The algorithm optionally includes a mixing step on line 5, where the rolled-out trajectories may replace the training trajectories with increasing probability approaching 1, which is similar to scheduled sampling \cite{bengio2015scheduled}, and may help stabilize learning in the early phase of the algorithm. In our main experiment, we do not notice a performance gain using this option.


\subsection{Joint Multi-Agent Imitation Learning}
\label{subsec:imitation_learning}

In this section we describe the $\texttt{Learn}$ procedure for multi-agent imitation learning  in Line 3 of Algorithm \ref{algo:main_algo}.
As background, for single agent imitation learning, reduction-based methods operate by iteratively collecting a new data set $\mathcal{D}_n$ at each round $n$ of training, consisting of state-action pairs $(s_t,a^*_t)$  where  $a^*_t$ is some optimal or demonstrated action given state $s_t$. A new policy can be formed by (i) combining a new policy from this data set $\mathcal{D}_n$ with previously learned policy $\pi$ \cite{daume2009search} or (ii) learning a new policy $\pi$ directly from the data set formed by aggregating $\mathcal{D}_1,\ldots,\mathcal{D}_n$ \cite{dagger}. Other variants exist although we do not discuss them here. 

The intuition behind the iterative reduction approach is to prevent a mismatch in training and prediction distributions due to sequential cascading errors (also called covariate-shift). The main idea is to use the learned policy's own predictions in the construction of subsequent states, thus simulating the test-time performance during training. This mechanism enables the agent to learn a policy that is robust to its own mistakes. Reduction-based methods also accommodate any black-box supervised training subroutine. We focus on using expressive function classes such as Long Short-Term Memory networks (LSTM) \citep{hochreiter1997long} as the policy class.\footnote{Note that conventional training of LSTMs does not address the cascading error problem.  While LSTMs are very good at sequence-to-sequence prediction tasks, they cannot naturally deal with the drifting of input state distribution drift caused by action output feedback in dynamical systems \cite{bengio2015scheduled}.}

\begin{algorithm}[tb]
	\begin{small}
	\caption{ Joint Multi-Agent Imitation Learning \\ \texttt{Learn}$(A_1,A_2,\ldots,A_K,C)$}
	\label{algo:joint_imitation}
	\begin{algorithmic}[1]
    	\REQUIRE Ordered actions $A_k = \{ a_{t,k} \}_{t=1}^T\enskip \forall k$, context $\{ c_t\}_{t=1}^T$
    	\REQUIRE Initialized policies $\pi_1,\ldots, \pi_K$
        \REQUIRE base routine $\train(S,A)$ mapping state to actions
		\STATE Set increasing prediction horizon $j\in\{ 1,\ldots,T\}$
		\FOR{$t = 0,j,2j,\ldots,T$}
		\FOR{$i=0,1,\ldots,j-1$} 
		\STATE Roll-out $\hat{a}_{t+i,k} = \pi_k(\hat{s}_{t+i-1,k})\quad \forall$ agent $k$ \\
		\STATE Cross-update for each policy $k\in \{ 1,\ldots,K\}$ \\ 
		$\hat{s}_{t+i,k} = \varphi_k \left( \left[\hat{a}_{t+i,1},\ldots, \hat{a}_{t+i,k},\ldots,\hat{a}_{t+i,K}, c_{t+i} \right] \right) $ \\
		\ENDFOR \\
		\STATE Policy update for all agent $k$ \\
		$\pi_k \leftarrow \train(\{\hat{s}_{t+i,k}, a_{t+i+1,k}^* \}_{i=0}^j) $
		\ENDFOR \\
		\OUTPUT $K$ updated policies $\pi_1,\pi_2,\ldots,\pi_K$
	\end{algorithmic}
	\end{small}
\end{algorithm}

Algorithm \ref{algo:joint_imitation} outlines the $\texttt{Learn}$ procedure for stable multi-agent imitation learning.
Assume we are given consistently indexed demonstrations $A = \left[ A_1,\ldots, A_K\right]$, where each $A_k = \{ a_{t,k}\}_{t=1}^T$ corresponds action of policy $\pi_k$. 
Let the corresponding expert action be $a_{t,k}^*$.
To lighten the notation, we denote the per-agent state vector by $s_{t,k} = \varphi_k([a_{t,1},\ldots,a_{t,k},\ldots,a_{t,K},c_t]) \footnote{Generally, state vector $s_{t,k}$ of policy $\pi_k$ at time $t$ can be constructed as $s_{t,k}=\left[\phi_k([a_{1:t,1}, c_{1:t}]),\ldots,\phi_k([a_{1:t,K}, c_{1:t}]) \right]$}$

Algorithm \ref{algo:joint_imitation} employs a roll-out horizon $j$, which divides the entire trajectory into $T/j$ segments. 
The following happens for every segment:
\vspace{-0.1in}
\begin{itemize}
\item Iteratively perform roll-out at each time step $i$ for all $K$ policies (line 4) to obtain actions $\{ \widehat{a}_{i,k}\}$.
\vspace{-0.05in}
\item Each policy simultaneously updates its state $\widehat{s}_{i,k}$, using the prediction from all other policies (line 5). 
\vspace{-0.05in}
\item At the end of the current segment, all policies are updated using the error signal from the deviation between predicted $\widehat{a}_{i,k}$ versus expert action $a^*_{i,k}$, for all $i$ along the sub-segment (line 7). 
\end{itemize} 
\vspace{-0.1in}
After policy updates, the training moves on to the next $j$-length sub-segment, using the freshly updated policies for subsequent roll-outs. The iteration proceeds until the end of the sequence is reached. In the outer loop the roll-out horizon $j$ is incremented.

Two key insights behind our approach are:
\vspace{-0.1in}
\begin{itemize}
	\item In addition to the training-prediction mismatch issue in single-agent learning, each agent's prediction must also be robust to imperfect predictions from  other agents. This non-stationarity issue also arises in multi-agent reinforcement learning \cite{busoniu2008comprehensive} when agents learn simultaneously. We perform joint training by cross-updating each agent's state using previous predictions from other agents. 
\vspace{-0.05in}
	\item Many single-agent imitation learning algorithms assume the presence of a dynamic oracle to provide one-step corrections $a^*_t$ along the roll-out trajectories. In practice, dynamic oracle feedback is very expensive to obtain and some recent work have attempted to relax this requirement \cite{le2016smooth,ho2016generative}. Without dynamic oracles, the rolled-out trajectory can deviate significantly from demonstrated trajectories when the prediction horizon $j$ is large ($\approx T$), leading to training instability. Thus $j$ is gradually increased to allow for slowly learning to make good sequential predictions over longer horizons.
\end{itemize}
\vspace{-0.1in}

For efficient training, we focus on stochastic optimization, which can invoke base routine $\texttt{Train}$ multiple times and thus naturally accommodates varying $j$. Note that the batch-training alternatives to Algorithm \ref{algo:joint_imitation} can also employ similar training schemes, with similar theoretical guarantees lifted to the multi-agent case. The Appendix shows how to use DAgger \cite{dagger} for Algorithm \ref{algo:joint_imitation}, which we used for our synthetic experiment. 


\subsection{Coordination Structure Learning} 
\label{subsec:svi}
The coordination mechanism is based on a latent structured model that governs the role assignment. The training and inference procedures seek to address two main issues: 
\vspace{-0.1in}
\begin{itemize}
	\item $\texttt{LearnStructure}$: unsupervised learning a probabilistic role assignment model $q$.
	\vspace{-0.05in}
	\item $\texttt{Assign}$: how $q$ informs the indexing mechanism so that unstructured trajectories can be mapped to structured trajectories amenable to Algorithm \ref{algo:joint_imitation}.
\end{itemize}
\vspace{-0.1in}
Given an arbitrarily ordered set of trajectories $U = \{U_1,\ldots,U_K,C \}$, let the coordination mechanism underlying each such $U$ be governed by a true unknown model $p$, with global parameters $\theta$. We suppress the agent/policy subscript and consider a generic featurized trajectory $x_{t} = [u_{t},c_t]\enskip \forall t$.  Let the latent role sequence for the same agent be $z = z_{1:T}$.
At any time $t$, each agent is acting according to a latent role $z_t\sim \texttt{Categorical}\{ \bar{1},\bar{2},\ldots,\bar{K}\}$, which are the local parameters to the structured model. 

Ideally, role and index asignment can be obtained by calculating the true posterior $p(z|x,\theta)$, which is often intractable.  We instead aim to approximate $p(z|x,\theta)$ by a simpler distribution $q$ via 
techniques from stochastic variational inference \cite{hoffman2013stochastic}, which allows for efficient stochastic training on mini-batches that can naturally integrate with our imitation learning subroutine. 

In variational inference, posterior approximation is often cast as optimizing over a simpler model class $\mathcal{Q}$, via searching for  parameters $\theta$ and $z$ that maximize the evidence lower bound (ELBO) $\mathcal{L}$:
\begin{equation*}
\mathcal{L}\left(q(z,\theta) \right) \triangleq \E_q \left[ \ln p(z,\theta,x)\right] - \E_q \left[ \ln q(z,\theta)\right] \leq \ln p(x)
\end{equation*}
Maximizing $\mathcal{L}$ is equivalent to finding $q\in\mathcal{Q}$ to minimize the KL divergence $\text{KL}\left( q(z,\theta |x) || p(z,\theta|x) \right)$.  We focus on the structured mean-field variational family, which factorizes $q$ as $q(z,\theta) = q(z)q(\theta)$. 
This factorization breaks the dependency between  $\theta$ and $z$, but not between single latent states $z_t$, unlike variational inference for i.i.d data \cite{kingma2013auto}.

\subsubsection{Training to learn model parameters}
The procedure to learn the parameter of our structured model is summarized in Algorithm \ref{algo:structure_learning}. Parameter learning proceeds via alternating updates over the factors $q(\theta)$ and $q(z)$, while keeping other factor fixed. Stochastic variational inference performs such updates efficiently in mini-batches. 
We slightly abuse notations and overload $\theta$ for the natural parameters of global parameter $\theta$ in the exponential family. Assuming the usual conjugacy in the exponential family, the stochastic \text{natural} gradient takes a convenient  form (line 2 of Algo \ref{algo:structure_learning}, and derivation in Appendix), where $t(z,x)$ is the vector of sufficient statistics, $b$ is a vector of scaling factors adjusting for the relative size of the mini-batches. Here the global update assumes optimal local update $q(z)$ has been computed. 

Fixing the global parameters, the local updates are based on message-passing over the underlying graphical model. The exact mathematical derivation depends on the specific graph structure. The simplest scenario is to assume independence among $z_t$'s, which resembles naive Bayes. In our experiments, we instead focus on Hidden Markov Models to capture first-order dependencies in role transitions over play sequences. In that case, line 1 of Algorithm \ref{algo:structure_learning} resembles running the forward-backward algorithm to compute the update $q(z)$. The forward-backward algorithm in the local update step takes $O(K^2T)$ time for a chain of length $T$ and $K$ hidden states. For completeness, derivation of parameter learning for HMMs is included in the Appendix. 

\begin{algorithm}[tb]
	\begin{small}
		\caption{ Structure Learning \\ \texttt{LearnStructure} 
			$\{U_{1},\ldots,U_{K},C,\theta,\rho \} \mapsto  q(\theta,z)$}
		\label{algo:structure_learning}
		\begin{algorithmic}[1]
        \REQUIRE $X_k = \{x_{t,k}\}_{t=1}^T = \{[u_{t,k},c_t]\} \enskip \forall t,k. X = \{ X_k\}_{k=1}^K$ \\
			Graphical model parameters $\theta$, stepsize $\rho$
			\STATE Local update: compute $q(z)$ via message-passing while fixing $\theta$ (See Appendix for derivations)
			\STATE Global parameter update: via natural gradient ascent \\
			$\theta \leftarrow \theta (1-\rho)+\rho(\theta_{prior}+b^\top \E_{q(z)}\left[ t(z,x)\right]) $
			\OUTPUT Updated model $q(\theta,z) = q(\theta)q(z)$
		\end{algorithmic}
	\end{small}
\end{algorithm}

\subsubsection{Inference for role and index assignment}

We can compute two types of inference on a learned $q$: 

\textbf{Role inference.} Compute the most likely role sequence $\{ z_{t,k}\}_{t=1}^T\in \{ \bar{1},\ldots,\bar{K}\}^T$, e.g., using  Viterbi (or dynamic programming-based forward message passing for graph structures). This most likely role sequence for agent $k$, which is the low-dimensional representation of the coordination mechanism, can be used to augment the contextual feature $\{ c_t\}_{t=1}^T$for each agent's policy training. 

\textbf{Role-based Index Assignment} Transform the unstructured set $U$ into an ordered set of trajectories $A$ to facilitate the imitation learning step.  This is the more important task for the overall approach.
The intuitive goal of an indexing mechanism is to facilitate consistent agent trajectory to policy mapping. 
Assume for notational convenience that we want index $k$ assigned to an unique agent who is most likely assuming role $\bar{k}$. Our inference technique rests on the well-known Linear Assignment Problem \cite{papadimitriou1982combinatorial}, which is solved optimally via the Kuhn-Munkres algorithm. Specifically, construct the cost matrix $M$ as:
\vspace{-0.05in}
\begin{align}
	M &= M_1\odot M_2 \label{eqn:cost_matrix} \\
	M_1 &= \left[ q(\{ x_{t,k}\} \vert z_{t,k}=  \bar{k})  \right]  = \left[\prod_{t=1}^T q(x_{t,k}|z_{t,k} = \bar{k}) \right] \nonumber  \\
	M_2 &= \left[ \ln q(\{ x_{t,k}\} \vert z_{t,k}  =  \bar{k}) \right] = \left[ \sum_{t=1}^T \ln q(x_{t,k}|z_{t,k} = \bar{k}) \right] \nonumber
\end{align}
where $k=1,\ldots,K, \bar{k}=\bar{1},\ldots,\bar{K},\odot$ is the Hadamard product, and matrices $M_1, M_2$ take advantage of the Markov property of the graphical model. 
Now solving the linear assignment problem for cost matrix $M$, we obtain the matching $\mathcal{A}$ from role $\bar{k}$ to index $k$, such that the total cost per agent is minimized. From here, we rearrange the unordered set $\{ U_1,\ldots, U_K\}$ to the ordered sequence $\left[ A_1,\ldots,A_K\right] \equiv [U_{\mathcal{A}(1)},\ldots,U_{\mathcal{A}(K)}]$ according to the minimum cost mapping. 

To see why this index assignment procedure results in an increased entropy in the original objective \eqref{eqn:objective}, notice that: 
\vspace{-0.1in}
\begin{align*}
	H(\mathcal{A} |\mathcal{D}) &\approx -\sum_{\bar{k}=1}^K P(\bar{k})q(\mathcal{A}(A_k) = \bar{k})\log q(\mathcal{A}(A_k) = \bar{k}) \\ 
	&=-\frac{1}{K}\sum_{\bar{k}=1}^K M(\bar{k},k),
\end{align*}
where we assume each latent role $\bar{k}$ has equal probability. The RHS increases from the linear assignment and consequent role assignment procedure. 
Our inference procedure to perform role assignment is summarized in Algorithm \ref{algo:assignment}.
\begin{algorithm}[tb]
	\begin{small}
		\caption{ Multi-Agent Role Assignment \\ \texttt{Assign} 
			$\{U_{1},\ldots,U_{K}\vert q \} \mapsto [A_1,\ldots,A_K] $}
		\label{algo:assignment}
		\begin{algorithmic}[1]
			\REQUIRE Approximate inference model $q$. Unordered trajectories $U = \{ U_{k}\}_{k=1}^K$.   \\
			\STATE Calculate cost matrix $M\in \mathbb{R}^{K\times K}$ per equation \ref{eqn:cost_matrix}
			\STATE $\mathcal{A}\leftarrow\texttt{MinCostAssignment}(M)$ \\
			\OUTPUT $A_k = U_{\mathcal{A}(k)}\quad \forall k=1,2,\ldots,K$
		\end{algorithmic}
	\end{small}
\end{algorithm}


%% file: sec-experiment.tex
\section{Experiments}
\label{experiment}

We present empirical results from two settings.  The first is a synthetic setting based on predator-prey, where the goal is to imitate a coordinating team of predators.  The second is a large-scale imitation learning setting from player trajectores in professional soccer games, where the goal is to imitate defensive team play.

\subsection{Predator-Prey Domain}
\textbf{Setting. }The predator-prey problem, also frequently called the Pursuit Domain \cite{benda1985optimal}, is a popular setting for multi-agent reinforcement learning. The traditional setup is with four predators and one prey, positioned on a grid board. 
At each time step, each agent has five moves: 
\begin{wrapfigure}{r}{0.21\textwidth}
		\vspace{-10pt}
	\centering
	\includegraphics[scale=0.2]{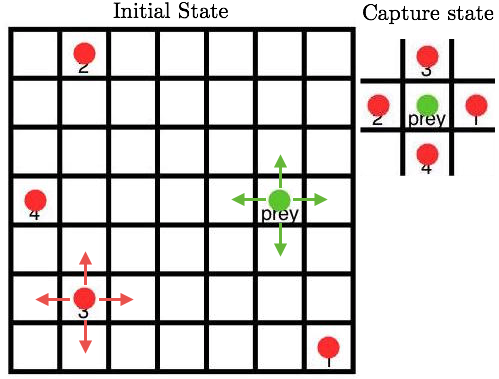}
	\vspace{-0.3in}
	\caption{
	\label{fig:predprey_interface}}
	\vspace{-10pt}
\end{wrapfigure}
N,S,E,W or no move. The world is toroidal:  the agents can move off one end of the board and come back on the other end. Agents make move simultaneously, but two agents cannot occupy the same position, and collisions are avoided by assigning a random move priority to the agents at each time step.  The predators can capture the prey only if the prey is surrounded by all four predators. The goal of the predators is to capture the prey as fast as possible, which necessarily requires coordination.  

\textbf{Data. }The demonstration data is collected from 1000 game instances, where four experts, indexed $1$ to $4$, are prescribed the consistent and coordinated role as illustrated in the capture state of Figure \ref{fig:predprey_interface}. In other words, agent $1$ would attempt to capture the prey on the right hand side, which allows for one fixed role for each expert throughout the game. However, the particular role assignment is hidden from the imitation learning task.
Each expert is then exhaustively trained using Value Iteration \cite{sutton1998reinforcement} in the reinforcement learning setting, with the reward of $1$ if the agent is in the position next to the prey according to its defined role, and $0$ otherwise. A separate set of 100 games was collected for evaluation. A game is terminated after 50 time steps if the predators fail to capture the prey. In the test set, the experts fail to capture the prey in $2\%$ of the games, and on average take $18.3$ steps to capture the prey.  

\textbf{Experiment Setup.} For this experiment, we use the batch version of Algorithm \ref{algo:main_algo} (see appendix) to learn to imitate the experts using only demonstrations.  Each policy is represented by a random forest of $20$ trees, and were trained over 10 iterations. The expert correction for each rolled-out state is collected via Value Iteration. The experts thus act as dynamic oracles, which result in a multi-agent training setting analogous to DAgger \cite{dagger}. We compare two versions of multi-agent imitation learning: 
\vspace{-0.1in}
\begin{itemize}
\item \textbf{Coordinated Training.} We use our algorithm, with the latent structure model represented by a discrete Hidden Markov Model with binomial emission. We use Algorithm \ref{algo:assignment} to maximize the role consistency of the dynamic oracles across different games.
\vspace{-0.05in}
\item \textbf{Unstructured Training.} An arbitrary role is assigned to each dynamic oracle for each game, i.e., the agent index is meaningless. 
\end{itemize}
\vspace{-0.1in}
In both versions, training was done using the same data aggregation scheme and batch training was conducted using the same random forests configuration. 

\begin{figure}[t]
	\centering
	\includegraphics[scale=0.42]{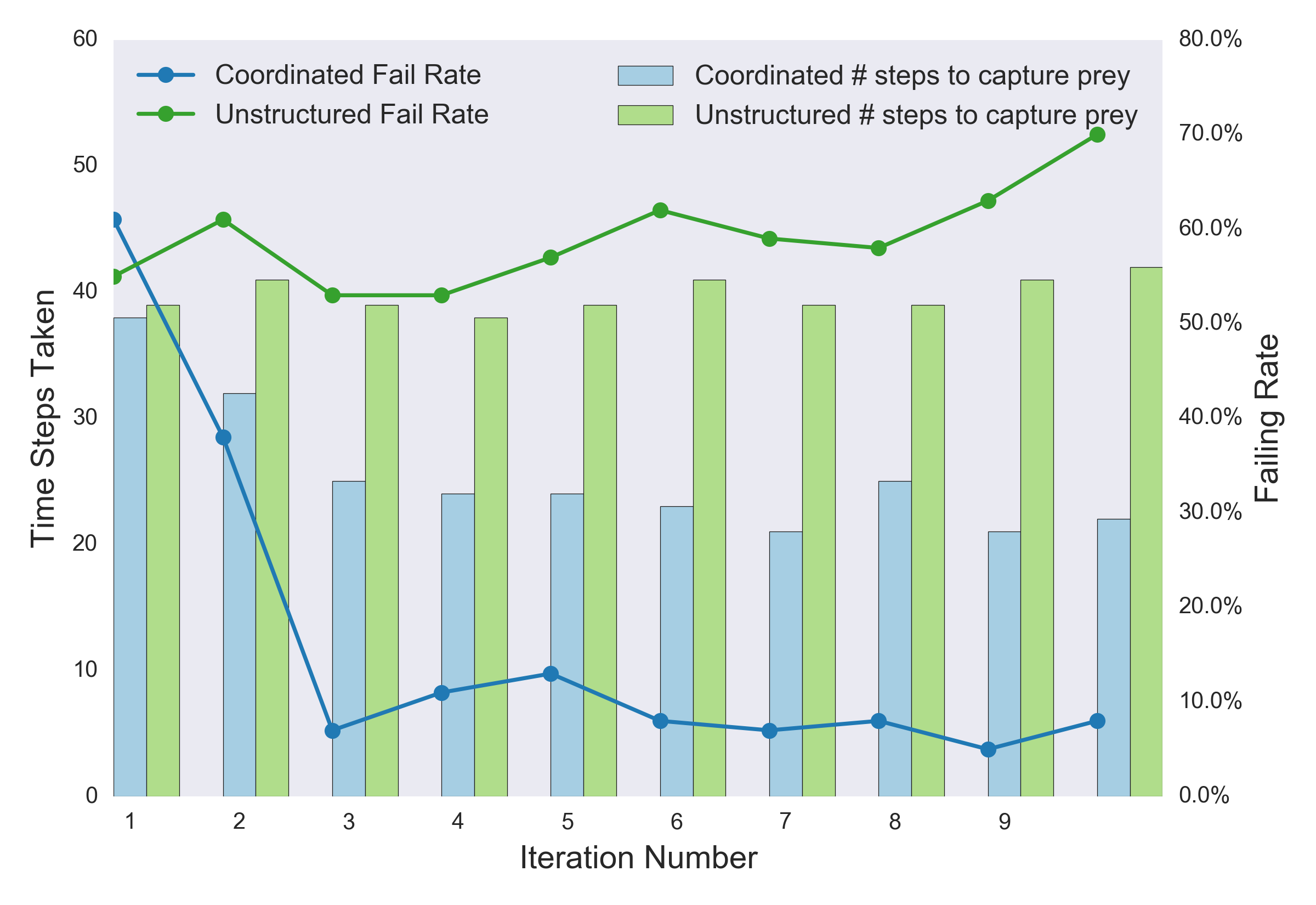}
   \vspace{-0.3in}
	\caption{\textit{Comparing performance in Predator-Prey between our approach and unstructured multi-agent imitation learning, as a function of the number of training rounds. Our approach demonstrates both significantly lower failure rates as well as lower average time to success (for successful trials).}
		\label{fig:pred_prey}}
\end{figure}

\textbf{Results. } Figure \ref{fig:pred_prey} compares the test performance of our method versus unstructured multi-agent imitation learning. 
Our method quickly approaches expert performance (average 22 steps with $8\%$ failure rate in the last iteration), whereas unstructured multi-agent imitation learning performance did not improve beyond the first iteration (average 42 steps with 70\% failure rate). Note that we even gave the unstructured baseline some advantage over our method, by forcing the prey to select the moves last after all predators make decisions (effectively making the prey slower). Without this advantage, the unstructured policies fail to capture the prey almost $100\%$ of the time. Also, if the same restriction is applied to the policies obtained from our method, performance would be on par with the experts ($100\%$ success rate, with similar number of steps taken). 

\subsection{Multi-agent Imitation Learning for Soccer}
\textbf{Setting. } Soccer is a popular domain for multi-agent learning. RoboCup, the robotic and simulation soccer platform, is perhaps the most popular testbed for multi-agent reinforcement learning research to date \cite{AAAI16-WhatsHot}. The success of MARL has been limited, however, due to the extremely high dimensionality of the problem. In this experiment, we aim to learn multi-agent policies for team soccer defense, based on tracking data from real-life professional soccer \cite{formation}. 

\textbf{Data. } We use the tracking data from 45 games of real professional soccer from a recent European league. The data was chunked into sequences with one team attacking and the other defending. Our goal is to learn up to 10 policies for team defense (11 players per team, minus the goal keeper). The training data consists of 7500 sets of trajectories $A = \{ A_1,\ldots,A_{10}\}$ , where $A_k = \{ a_{t,k} \}_{t=1}^T$ is the sequence of positions of one defensive player, and $C$ is the context consisting of opponents and the ball. Overall, there are about 1.3 million frames at 10 frames per second. The average sequence length is 176 steps, and the maximum is 1480.

\textbf{Experiment Setup. } Each policy is represented by a recurrent neural network structure (LSTM), with two hidden layers of $512$ units each. As LSTMs generally require fixed-length input sequences, we further chunk each trajectory into sub-sequences of length 50, with overlapping window of 25 time steps. The joint multi-agent imitation learning procedure follows Algorithm \ref{algo:joint_imitation} closely. In this set-up, without access to dynamic oracles for imitation learning in the style of SEARN \cite{daume2009search} and DAgger \cite{dagger}, we gradually increase the horizon of the rolled-out trajectories from 1 to $10$ steps look-ahead. Empirically, this has the effect of stabilizing the policy networks early in training, and limits the cascading errors caused by rolling-out to longer horizons.

\begin{figure}[t]
	\centering
	\includegraphics[scale=0.73]{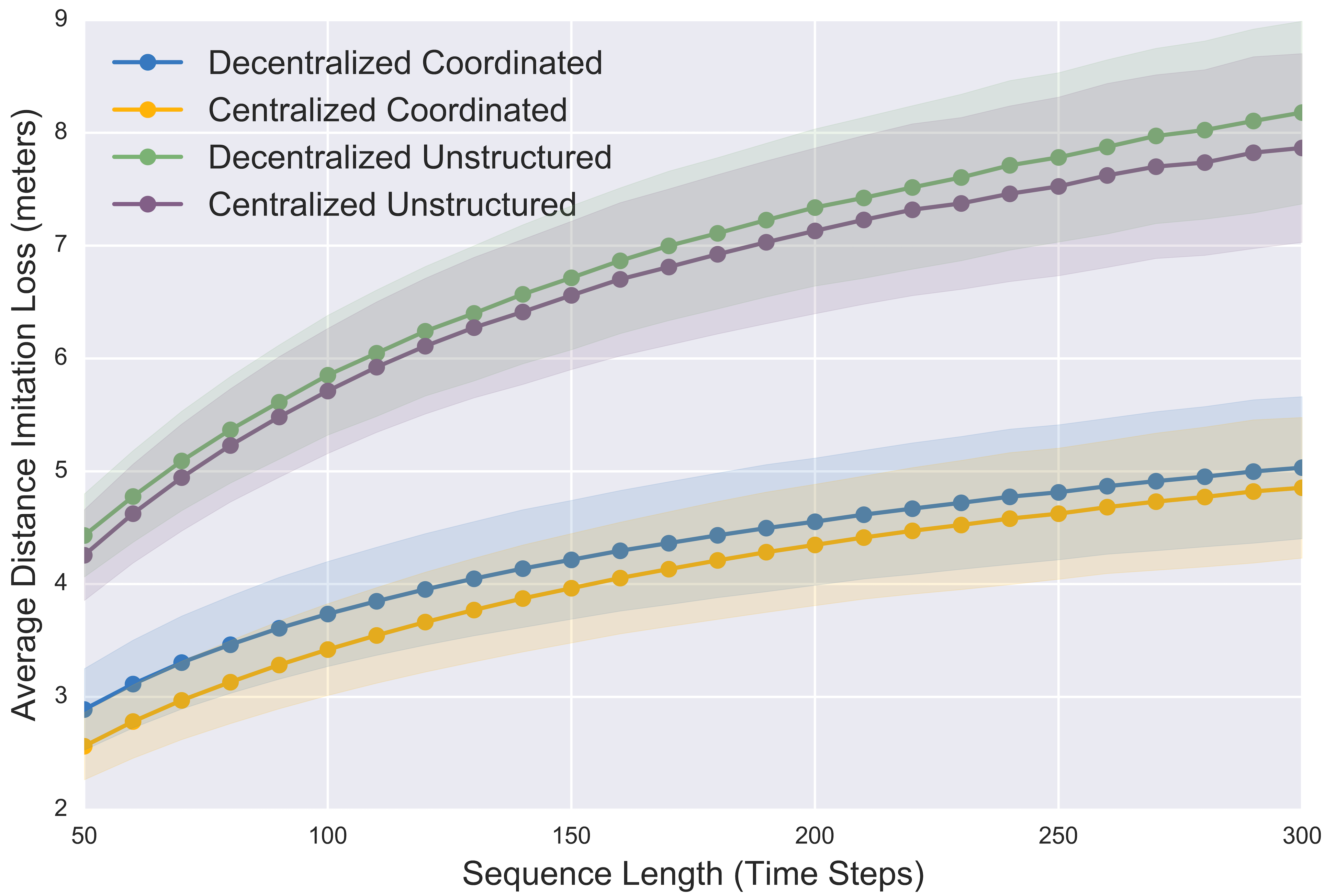}
	\vspace{-0.1in}
	\caption{\textit{Experimental results on soccer domain.   We see that using coordination substantially improves the imitation loss, and that the decentralized policy is comparable to the centralized.}
		\label{fig:experiment_compare}}
\end{figure}

The structured model component is learned via stochastic variational inference on a continuous HMM, where the per-state emission distribution is a mixture of Gaussians. Training and inference operate on the same mini-batches used for joint policy learning.  

We compare against two variations. The first employs centralized policy that aggregates the state vectors of all decentralized learner and produces the actions for all players, i.e., a multi-task policy. The centralized approach generally requires more model parameters, but is potentially much more accurate.
The second variation is to not employ joint multi-agent training:  we modify Algorithm \ref{algo:joint_imitation} to not cross-update states between agents, and each role is trained conditioned on the ground truth of the other agents.



\textbf{Results.}  Figure \ref{fig:experiment_compare} shows the results.  Our coordinated learning approach substantially outperforms conventional imitation learning without structured coordination.
The imitation loss measures average distance of roll-outs and ground truth in meters (note the typical size of soccer field is $110\times70$ meters). As expected, average loss increases with longer sequences, due to cascading errors. However, this error scales sub-linearly with the length of the horizon, even though the policies were trained on sequences of length 50. Note also that the performance difference between decentralized and centralized policies is insignificant compared to the gap between coordinated and unstructured policies, further highlighting the benefits of structured coordination in multi-agent settings. The loss of a single network, non-joint training scheme is very large and thus omitted from Figure \ref{fig:experiment_compare} (see the appendix). 

\begin{figure}[t]
	\centering
	\includegraphics[scale=0.5]{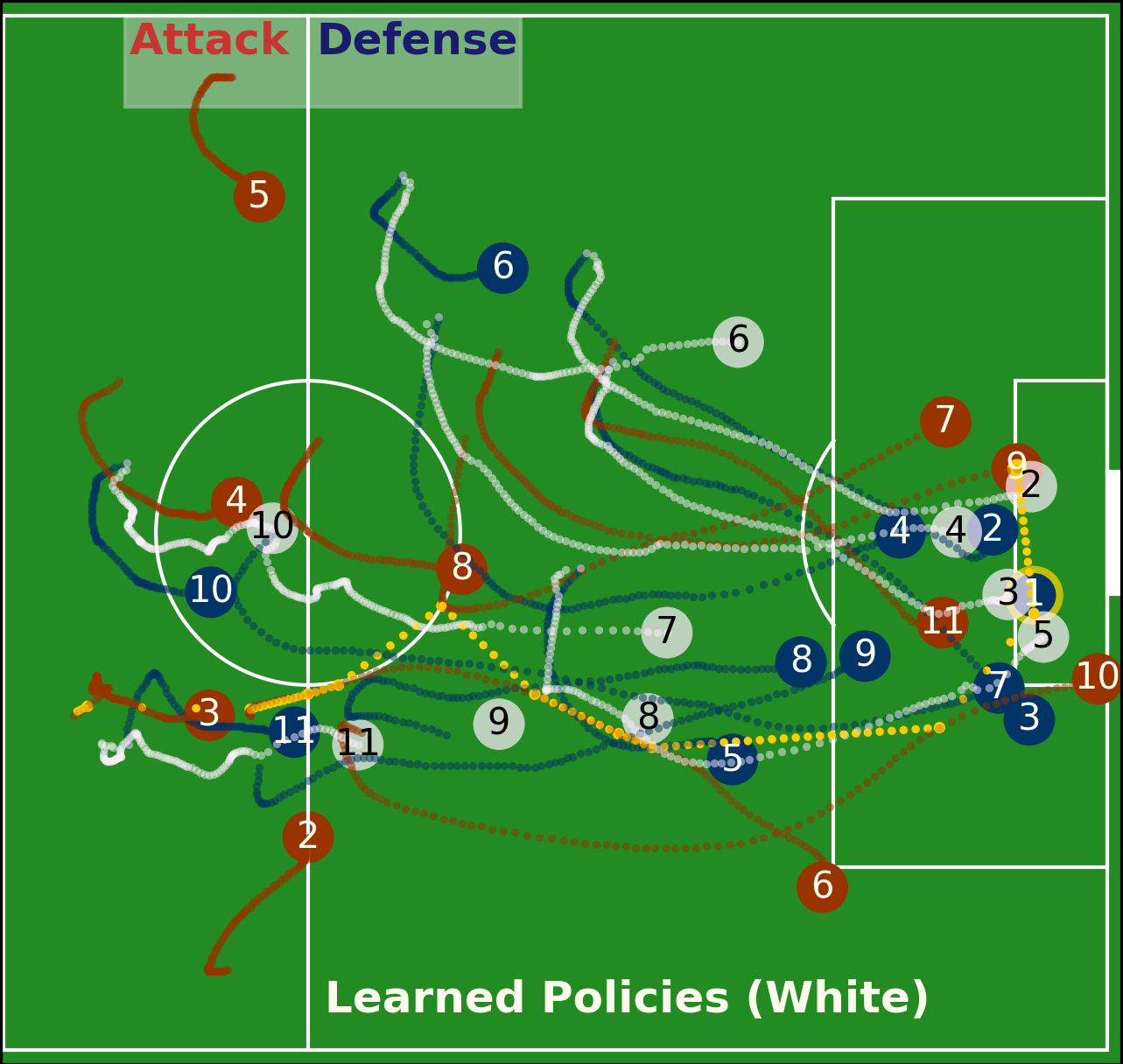}
  \vspace{-0.1in}
	\caption{\textit{Result of 10 coordinated imitation policies, corresponding with Figure \ref{fig:game_example}. White is the rolled-out imitation policies.}
	\label{fig:ghost_trajectory}}
\end{figure}

\begin{figure}[t]
	\centering
     \includegraphics[scale=0.45]{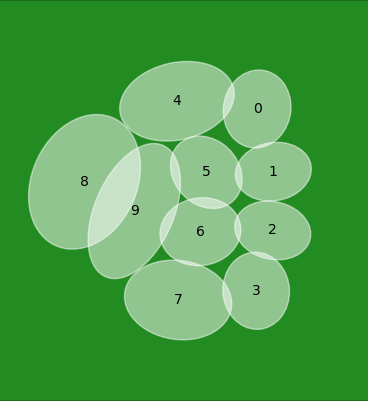}
	\vspace{-0.1in}
	\caption{\textit{Components of role distributions, corresponding to a popular formation arrangement in professional soccer}
		\label{fig:role_components}}
\end{figure}

\textbf{Visualizations.} Imitation loss, of course, is not a full reflection of the quality of the learned policies. Unlike predator-prey, the long-term reward signal is not available, so we rely on visual inspection as part of evaluation. Figure \ref{fig:ghost_trajectory} overlays policy prediction on top of the actual game sequence from Figure \ref{fig:game_example}. Additional test examples are included in our supplemental video \footnote{Watch video at \url{http://hoangminhle.github.io}}.  We note that learned policies are qualitatively similar to the ground truth demonstrations, and can be useful for applications such as counterfactual replay analysis \cite{sloan2017}. Figure \ref{fig:role_components} displays the Gaussian components of the underlying HMM. The components correspond to the dominant modes of the roles assigned. Unlike the predator-prey domain, roles can be switched during a sequence of play. See the appendix for more details on role swap frequency.

%% file: sec-related.tex
\section{Other Related Work}
\label{related}
The problem of multi-agent imitation learning has not been widely considered, perhaps with the exception of \cite{chernova2007multiagent} which focused on very different applications and technical challenges (i.e., learning a model of a joint task by collecting samples from direct interaction with teleoperating human teachers). The actual learning algorithm there requires the learner to collect enough data points from human teachers for confident classification of task. It is not clear how well the proposed method would translate to other domains.

Index-free policy learning is generally difficult for black-box machine learning techniques. Some recent work has called attention to the importance of order to learning when input or output are sets \cite{vinyals2015order}, motivated by classic algorithmic and geometric problems such as learning to sort a set of numbers, or finding convex hull for a set of points, where no clear indexing mechanism exists. 
Other permutation invariant approaches include those for standard classification \cite{shivaswamy2006permutation}.

%% file: sec-conclude.tex
\section{Limitations and Future Work}
\label{conclusion}
In principle, the training and inference of the latent structure model can accommodate different types of graphical models. However, the exact procedure varies depending on the graph structure. 
It would be interesting to find domains that can benefit from more general graphical models. Another possible direction is to  develop fully end-to-end differentiable training methods that can accommodate our index-free policy learning formulation, especially deep learning-based method that could provide computational speed-up compared to traditional graphical model inference. One potential issue with the end-to-end approach is the need to depart from a learning-reductions style approach.

Although we addressed learning from demonstrations in this paper, the proposed framework can also be employed for generative modeling, or more efficient structured exploration for reinforcement learning. Along that line, our proposed method could serve as a useful component of general reinforcement learning, especially in multi-agent settings where traditional exploration-based approaches such as Q-learning prove computationally intractable.

%% file: sec-derivation.tex
\section{Variational Inference Derivation for Hidden Markov Models}
In this section, we provide the mathematical derivation for the structured variational inference procedure. We focus on the training for Bayesian Hidden Markov Model, in particular the Forward-Backward procedure to complete the description of Algorithm \ref{algo:structure_learning}. The mathematical details for other types of graphical models depend on the family of such models and should follow similar derivations. Further relevant details on stochastic variational inference can be found in \cite{hoffman2013stochastic, johnson2014stochastic, beal2003variational}. d

\textbf{Settings.} Given an arbitrarily ordered set of trajectories $U = \{U_1,\ldots,U_K,C \}$, let the coordination mechanism underlying each such $U$ be governed by a true unknown model $p$, with global parameters $\theta$. We suppress the agent/policy subscript and consider a generic featurized trajectory $x_{t} = [u_{t},c_t]\enskip \forall t$.  Let the latent role sequence for the same agent be $z = z_{1:T}$.

At any time $t$, each agent is acting according to a latent role $z_t\sim \texttt{Categorical}\{ \bar{1},\bar{2},\ldots,\bar{K}\}$, which are the local parameters to the structured model. 

Ideally, role and index asignment can be obtained by calculating the posterior $p(z|x,\theta)$, which is often intractable.  One way to infer the role assignment is via approximating the intractable posterior $p(z|x,\theta)$ using Bayesian inference, typically via MCMC or mean-field variational methods. Since sampling-based MCMC methods are often slow, we instead aim to learn to approximate $p(z|x,\theta)$ by a simpler distribution $q$ via Bayesian inference. In particular, we employ techniques from stochastic variational inference \cite{hoffman2013stochastic}, which allows for efficient stochastic training on mini-batches that can naturally integrate with our imitation learning subroutine. 

\textbf{Structured Variational Inference for Unsupervised Role Learning.} Consider a full probabilistic model:
\begin{equation*}
p(\theta, z, x) = p(\theta)\prod_{t=1}^T p(z_t|\theta)p(x_t|z_t,\theta)
\end{equation*}
with global latent variables $\theta$, local latent variables $z = \{ z_t\}_{t=1}^T$.
Posterior approximation is often cast as optimizing over a simpler model class $\mathcal{Q}$, via searching for global parameters $\theta$ and local latent variables $z$ that maximize the evidence lower bound (ELBO) $\mathcal{L}$:
\begin{align}
	\log p(x) &\geq \E_q \left[ \log p(z,\theta,x)\right] - \E_q \left[ \log q(z,\theta)\right]  \nonumber \\
	&\triangleq \mathcal{L}\left(q(z,\theta) \right). \nonumber
\end{align}
Maximizing $\mathcal{L}$ is equivalent to finding $q\in\mathcal{Q}$ to minimize the KL divergence $\text{KL}\left( q(z,\theta |x) || p(z,\theta|x) \right)$.  

For unsupervised structured prediction problem over a family of graphical model, we focus on the structured mean-field variational family, which factorizes $q$ as $q(z,\theta) = q(z)q(\theta)$ \cite{hoffman2014structured} and decomposes the ELBO objective:
\begin{align}
	\mathcal{L} &= \E_q[\log p(\theta]-\E_q[\log q(\theta] \nonumber \\ 
	& \quad +\E_q[\log(p(z,x|\theta)]-\E_q[\log(q(z))].
\end{align}
This factorization breaks the dependency between  $\theta$ and $z$, but not between single latent states $z_t$, unlike variational inference for i.i.d data \cite{kingma2013auto}.

Variational inference optimizes the objective $\mathcal{L}$ typically using natural gradient ascent over global factors $q(\theta)$ and local factors $q(z)$. (Under mean-field assumption, optimization typically proceeds via alternating updates of $\theta$ and $z$.) Stochastic variational inference performs such updates efficiently in mini-batches. For graphical models, structured stochastic variational inference optimizes $\mathcal{L}$ using \textit{natural gradient} ascent over global factors $q(\theta)$ and message-passing scheme over local factors $q(z)$. We assume the prior $p(\theta)$ and complete conditionals $p(z_t,x_t|\theta)$ are conjugate pairs of exponential family, which gives natural gradient of $\mathcal{L}$ with respect to $q(\theta)$ convenient forms \cite{johnson2014stochastic}. Denote the exponential family forms of $p(\theta)$ and $p(z_t,y_t | \theta)$ by:
\begin{align*}
\ln p(\theta) &= \langle\eta_\theta, t_{\theta}(\theta) \rangle - A_{\theta}(\eta_{\theta}) \\ 
\ln p(z_t,x_t | \theta) &= \langle \eta_{zx}(\theta), t_{zx}(z_t,x_t) \rangle - A_{zx}(\eta_{zx}(\theta))
\end{align*}
where $\eta_\theta$ and $\eta_{zx}$ are functions indicating natural parameters, $t_\theta$ and $t_{zx}$ are sufficient statistics and $A(\cdot)$ are log-normalizers (\cite{blei2017variational}). Note that in general, different subscripts corresponding to $\eta, t, A$ indicate different function parameterization (not simply a change in variable value assignment). Conjugacy in the exponential family yields that \cite{blei2017variational}:
\begin{equation*}
t_{\theta}(\theta) = \left[ \eta_{zx}(\theta), -A_{zx}(\eta_{zx}(\theta))\right]
\end{equation*}
and that
\begin{equation}
\label{eqn:conjugacy_form}
p(\theta|z_t,x_t) \propto \exp \{ \langle \eta_{\theta}+\left[ t_{zx}(z_t,x_t),1 \right] , t_{\theta}(\theta) \rangle \}
\end{equation}
Conjugacy in the exponential family also implies that the optimal $q(\theta)$ is in the same family \cite{blei2017variational},  i.e.
\begin{equation*}
q(\theta) = \exp\{ \langle \widetilde{\eta}_{\theta}, t_{\theta}(\theta) \rangle - A_{\theta}(\widetilde{\eta}_{\theta}) \}
\end{equation*}
for some natural parameters $\widetilde{\eta}_{\theta}$ of $q(\theta)$. 

To optimize over global parameters $q(\theta)$, conjugacy in the exponential family allows obtaining convenient expression for the gradient of $\mathcal{L}$ with respect to natural parameters $\widetilde{\eta}_{\theta}$. The derivation is shown similarly to \cite{johnson2014stochastic} and \cite{blei2017variational} - we use simplified notations $\widetilde{\eta}\triangleq \widetilde{\eta}_{\theta}, \eta \triangleq \eta_{\theta}, A \triangleq A_{\theta}$, and $t(z,x)\triangleq \sum_{t=1}^T \left[ t_{zx}(z_t,x_t),1 \right]$. Taking advantage of the exponential family identity $\mathbb{E}_{q(\theta)}[t_{\theta}(\theta)] = \nabla A(\widetilde{\eta})$, the objective $\mathcal{L}$ can be re-written as:
\begin{align*}
\mathcal{L} &= \mathbb{E}_{q(\theta)q(z)} \left[ \ln p(\theta |z,x) - \ln q(\theta) \right] \\ 
&= \langle\eta + \mathbb{E}_{q(z)}[t(z,x)], \nabla A(\widetilde{\eta}) \rangle- (\langle \widetilde{\eta},\nabla A(\widetilde{\eta} ) \rangle - A(\widetilde{\eta}))
\end{align*}
Differentiating with respect to $\widetilde{\eta}$, we have that
\begin{equation*}
\nabla_{\widetilde{\eta}} \mathcal{L} = \left(\nabla^2 A(\widetilde{\eta}) \right) \left(\eta + \mathbb{E}_{q(z)}[t(z,x)] - \widetilde{\eta} \right)
\end{equation*}
The \textit{natural gradient} of $\mathcal{L}$, denoted $\widetilde{\nabla}_{\widetilde{\eta}}$, is defined as $\widetilde{\nabla}_{\widetilde{\eta}} \triangleq \left(\nabla^2 A(\widetilde{\eta}) \right)^{-1} \nabla_{\widetilde{\eta}}$. And so the natural gradient of $\mathcal{L}$ can be compactly described as:
\begin{equation}
\label{natural_gradient_form}
\widetilde{\nabla}_{\widetilde{\eta}} \mathcal{L} = \eta + \sum_{t=1}^T \mathbb{E}_{q(z_t)}\{ [t_{zx}(z_t,x_t), 1] \} - \widetilde{\eta}
\end{equation}
Thus a stochastic natural descent update on the global parameters $\widetilde{\eta}_{\theta}$ proceeds at step $n$ by sampling a mini-batch $x_t$ and taking the global update with step size $\rho_n$:
\begin{equation}
\label{global_update}
\widetilde{\eta}_{\theta} \leftarrow (1-\rho_n)\widetilde{\eta}_{\theta} + \rho_n (\eta_\theta+b^\top \mathbb{E}_{q^*(z_t)}[t(z_t,x_t)] )
\end{equation}
where $b$ is a vector of scaling factors adjusting for the relative size of the mini-batches. Here the global update assumes optimal local update $q^*(z)$ has been computed. In each step however, the local factors $q^*(z_t)$ are computed with mean field updates and the current value of $q(\theta)$ (analogous to coordinate ascent). In what follows, we provide the derivation for the update rules for Hidden Markov Models, which are the particular instantiation of the graphical model we use to represent the role transition for our multi-agent settings. 

\textbf{Variational factor updates via message passing for Hidden Markov Models.}
For HMMs, we can view global parameters $\theta$ as the parameters of the underlying HMMs such as transition matrix and emission probabilities, while local parameters $z$ govern hidden state assignment at each time step. 

Fixing the global parameters, the local updates are based on message passing over the graphical model. The exact mathematical derivation depends on the specific graph structure. The simplest scenario is to assume independence among $z_t$'s, which resembles naive Bayes. We instead focus on Hidden Markov Models to capture first-order dependencies in role transitions over play sequences. In this case, global parameters $\theta = \left( p_0, P, \phi \right)$ where $P = \left[ P_{ij} \right]_{i,j=1}^K$ is the transition matrix with $P_{ij} = p(z_t = j|z_{t-1} = i)$, $\phi = \{ \phi_i\}_{i=1}^K$ are the emission parameters, and $p_0$ is the initial distribution.

Consider a Bayesian HMM on $K$ latent states. Priors on the model parameters include the initial state distribution $p_0$, transition matrix $P$ with rows denoted $p_1,\ldots, p_K$, and the emission parameters $\phi = \{ \phi_i \}_{i=1}^K$. In this case we have the global parameters $\theta = (p_0,P,\phi)$. For Hidden Markov Model with observation $x_{1:T}$ and latent sequence $z_{1:T}$, the generative model over the parameters is given by $\phi_i \sim p(\phi)$ (i.i.d from prior), $p_i \sim \text{Dir}(\alpha_i), z_1 \sim p_0, z_{t+1} \sim p_{z_t}$, and $x_t \sim p(x_t | \phi_{z_t})$ (conditional distribution given parameters $\phi$). 
 We can also write the transition matrix:
\[
P = 
\begin{bmatrix}
p_1 \\ \vdots \\p_K
\end{bmatrix}
\]
The Bayesian hierarchical model over the parameters, hidden state sequence $z_{1:T}$, and observation sequence $y_{1:T}$ is
\begin{align*}
\phi_i &\overset{\text{iid}}{\sim} p(\phi), p_i \sim \text{Dir}(\alpha_{i}) \\
z_1 &\sim p_0, z_{t+1} \sim p_{z_t}, x_t \sim p(x_t | \phi_{z_t})
\end{align*}

For HMMs, we have a full probabilistic model: $p(z,x | \theta) = p_0(z_1)\prod_{t=1}^T p(z_t | z_{t-1}, P) p(x_t|z_t,\phi)$. Define the likelihood potential $L_{t,i} = p(x_t |\phi_i)$, the likelihood of the latent sequence, given observation and model parameters, is as follows:
\begin{align}
& p(z_{1:T} | x_{1:T}, P, \phi) = \nonumber \\  
& \exp\left(\log p_0(z_1)+\sum_{t=2}^T \log P_{z_{t-1},z_t} + \sum_{t=1}^T\log L_{t, z_t} - Z\right) \label{eqn:likelihood_potential}
\end{align}
where $Z$ is the normalizing constant. 
Following the notation and derivation from \cite{johnson2014stochastic}, we denote $p(z_{1:T |x_{1:T}, P, \phi}) = \text{HMM}(p_0, P, L)$. Under mean field assumption, we approximate the true posterior $p(P,\phi, z_{1:T} | x_{1:T})$ with a mean field variational family $q(P)q(\phi)q(z_{1:T})$ and update each variational factor in turn while fixing the others. 

Fixing the global parameters $\theta$, taking expectation of log of (\ref{eqn:likelihood_potential}), we derive the update rule for $q(z)$ as $q(z_{1:T}) = \text{HMM}(\widetilde{P}, \widetilde{p}_0, \widetilde{L})$ where:
\begin{align*}
\widetilde{P}_{j,k} &= \exp \{ \E_{q(P)} \ln(P_{j,k}) \} \\
\widetilde{p}_{0,k} &= \exp\{ \ln \mathbb{E}_{q(p_0)}p_{0,k} \} \\ 
\widetilde{L}_{t,k} &= \exp\{ \E_{q(\phi_k)} \ln (p(x_t |z_t = k)) \}
\end{align*}
To calculate the expectation with respect to $q(z_{1:T})$, which is necessary for updating other factors, the $\texttt{Forward-Backward}$ recursion of HMMs is defined by forward messages $F$ and backward messages $B$:
\begin{align}
F_{t,i} &= \sum_{j=1}^K F_{t-1,j} \widetilde{P}_{j,i} \widetilde{L}_{t,i} \label{eqn:forward_pass} \\
B_{t,i} &= \sum_{j=1}^K \widetilde{P}_{i,j}\widetilde{L}_{t+1,j} B_{t+1,j} \label{eqn:backward_pass} \\
F_{1,i} &= p_0(i) \nonumber \\
B_{T,i} &= 1 \nonumber
\end{align}

As a summary, calculating the gradient w.r.t $z$ yields the following optimal variational distribution over the latent sequence:
\vspace{-0.1in}
\begin{align}
	q^*(z) &\propto \exp \Big(  \E_{q(P)}[\ln p_0(z_1)]+\sum_{t=2}^T\E_{q(P)}[\log P_{z_{t-1},z_t}] \nonumber \\ 
	&+ \sum_{t=1}^{T} \E_{q(\phi)}\ln [p(x_t |z_t)] \Big),
\end{align}
which gives the local updates for $q^*(z)$, given current estimates of $P$ and $\phi$:
\begin{align}
	\widetilde{P}_{j,k} &= \exp\left[ \E_{q(P)}\ln(P_{j,k})\right] \label{eqn:trans_matrix_update}\\
	\widetilde{p}(x_t | z_t = k) &= \exp\left[ E_{q(\phi)}\ln p(x_t|x_t=k) \right], \label{eqn:emission_update}
\end{align}
for $k=1,\ldots,K$, $t=1,\ldots,T$, and then use $p_0,\widetilde{P},\widetilde{p}$ to run the forward-backward algorithm to compute the update $q^*(z_t = k)$ and $q^*(z_{t-1} = j, z_t=k)$. The forward-backward algorithm in the local update step takes $O(K^2T)$ time for a chain of length $T$ and $K$ hidden states. 

\textbf{Training to learn model parameters for HMMs.} Combining \textit{natural gradient} step with message-passing scheme for HMMs yield specific update rules for learning the model parameters. Again for HMMs, the global parameters are $\theta = (p_0, P, \phi)$ and local variables $z = z_{1:T}$. 
Assuming the priors on observation parameter $p(\phi_i)$ and likelihoods $p(x_t|\phi_i)$ are conjugate pairs of exponential family distribution for all $i$, the conditionals $p(\phi_i | x)$ have the form as seen from equation \ref{eqn:conjugacy_form}:
\begin{equation*}
p(\phi_i|x) \propto \exp\{\langle \eta_{\phi_i}+[t_{x,i}(x),1], t_{\phi_i}(\phi_i) \rangle \}
\end{equation*}
For structured mean field inference, the approximation $q(\theta)$ factorizes as $q(P)q(p_0)q(\phi)$. At each iteration, stochastic variational inference sample a sequence $x_{1:T}$ from the data set (e.g. trajectory from any randomly sampled player) and perform stochastic gradient step on $q(P)q(p_0)q(\phi)$. In order to compute the gradient, we need to calculate expected sufficient statistics w.r.t the optimal factor for $q(z_{1:T})$, which in turns depends on current value of $q(P)q(p_0)q(\phi)$. 

Following the notation from \cite{johnson2014stochastic}, we write the prior and mean field factors as
\begin{align*}
p(p_i) &= \text{Dir}(\alpha_i), p(\phi_i) \propto \exp\{ \langle \eta_{\phi_i} , t_{\phi_i}(\phi_i) \rangle \} \\
q(p_i) &= \text{Dir}(\widetilde{\alpha}_i), q(\phi_i) \propto \exp\{ \langle \widetilde{\eta}_{\phi_i}, t_{\phi_i}(\phi_i) \rangle \}
\end{align*}
Using message passing scheme as per equations (\ref{eqn:forward_pass}) and (\ref{eqn:backward_pass}), we define the intermediate quantities:
\begin{align}
\widehat{t}_{x,i} &\triangleq \mathbb{E}_{q(z_{1:T})}\sum_{t=1}^T\mathbb{I}[z_t=i]t_{x,i}(x_t) \nonumber \\
&= \sum_{t=1}^T F_{t,i}B_{t,i} [t_{x,i}(x_t),1] / Z \\
(\widehat{t}_{trans,i})_j &\triangleq \mathbb{E}_{q(z_{1:T})}\sum_{t=1}^{T-1}\mathbb{I}[z_t = i, z_{t+1} = j] \nonumber \\ 
&= \sum_{t=1}^{T-1} F_{t,i}\widetilde{P}_{i,j}\widetilde{L}_{t+1,j}B_{t+1,j} /Z \\
(\widehat{t}_{init})_i &\triangleq \mathbb{E}_{q(z_{1:T})}\mathbb{I}[z_1=i]=\widetilde{p}_0 B_{1,i} /Z 
\end{align}
where $Z\triangleq \sum_{i=1}^K F_{T,i}$ is the normalizing constant, and $\mathbb{I}$ is the indicator function. 

Given these expected sufficient statistics, the specific update rules corresponding to the natural gradient step in the natural parameters of $q(P), q(p_0)$, and $q(\phi)$ become:
\begin{align}
\widetilde{\eta}_{\phi,i} &\leftarrow (1-\rho)\widetilde{\eta}_{\phi,i} + \rho (\eta_{\phi,i} + b^\top \widehat{t}_{x,i}) \label{eqn:phi_update}\\
\widetilde{\alpha}_{i} &\leftarrow (1-\rho)\widetilde{\alpha}_i+\rho(\alpha_i+b^\top \widehat{t}_{trans,i}) \label{eqn:trans_param_update} \\
\widetilde{\alpha}_0 &\leftarrow (1-\rho)\widetilde{\alpha}_0+\rho(\alpha_0 + b^\top \widehat{t}_{init,i}) \label{eqn:init_param_update}
\end{align}

\begin{algorithm}[tb]
	\begin{small}
		\caption{ Coordinated Structure Learning \\ \texttt{LearnStructure} 
			$\{U_{1},\ldots,U_{K},C,\theta,\rho \} \mapsto  q(\theta,z)$}
		\label{algo:structure_learning_detailed}
		\begin{algorithmic}[1]
			\REQUIRE Set of trajectories $U = \{ U_{k}\}_{k=1}^K$. Context $C$   \\
			Previous parameters $\theta = (p_0,\theta^P, \theta^\phi)$, stepsize $\rho$
			\STATE $X_k = \{x_{t,k}\}_{t=1}^T = \{[u_{t,k},c_t]\} \enskip \forall t,k. X = \{ X_k\}_{k=1}^K$
			\STATE Local update: Compute $\widetilde{P}$ and $\widetilde{p}$ per equation \ref{eqn:trans_matrix_update} and \ref{eqn:emission_update} \\
			and compute $q(z) = \texttt{Forward-Backward}(X,\widetilde{P},\widetilde{p})$
			\STATE Global update of $\theta$, per equations \ref{eqn:phi_update}, \ref{eqn:trans_param_update}, and \ref{eqn:init_param_update}. \\
			\OUTPUT Updated model $q(\theta,z) = q(\theta)q(z)$
		\end{algorithmic}
	\end{small}
\end{algorithm}



\section{Experimental Evaluation}
\subsection{Batch-Version of Algorithm \ref{algo:joint_imitation} for Predator-Prey}

\begin{algorithm}[h]
		\caption{ Multi-Agent Data Aggregation Imitation Learning \\ \texttt{Learn}$(A_1,A_2,\ldots,A_K,C | D)$}
		\label{algo:joint_dagger}
		\begin{algorithmic}[1]
			\REQUIRE Ordered actions $A_k = \{ a_{t,k} \}_{t=1}^T\enskip \forall k$, context $\{ c_t\}_{t=1}^T$
			\REQUIRE Aggregating data set $D_1,..,D_K$ for each policy
			\REQUIRE base routine $\train(S,A)$ mapping state to actions
			\FOR{$t = 0,1,2,\ldots,T$}
			\STATE Roll-out $\hat{a}_{t+1,k} = \pi_k(\hat{s}_{t,k})\quad \forall$ agent $k$ \\
			\STATE Cross-update for each policy $k\in \{ 1,\ldots,K\}$ \\ 
			$\hat{s}_{t+1,k} = \varphi_k \left( \left[\hat{a}_{t+1,1},\ldots, \hat{a}_{t+1,k},\ldots,\hat{a}_{t+1,K}, c_{t+1} \right] \right) $ \\
			\STATE Collect expert action $a_{t+1,k}^*$ given state $\hat{s}_{t+1,k} \enskip \forall k$ 
            \STATE Aggregate data set $D_k = D_k \cup \{\hat{s}_{t+1,k}, a_{t+1,k}^* \}_{t=0}^{T-1}$
			\ENDFOR \\
            \STATE $\pi_k \leftarrow \train(D_k) $
			\OUTPUT $K$ new policies $\pi_1,\pi_2,\ldots,\pi_K$
		\end{algorithmic}
\end{algorithm}

\subsection{Visualizing Role Assignment for Soccer}
\begin{figure}[h]
	\centering
	\includegraphics[scale=0.3]{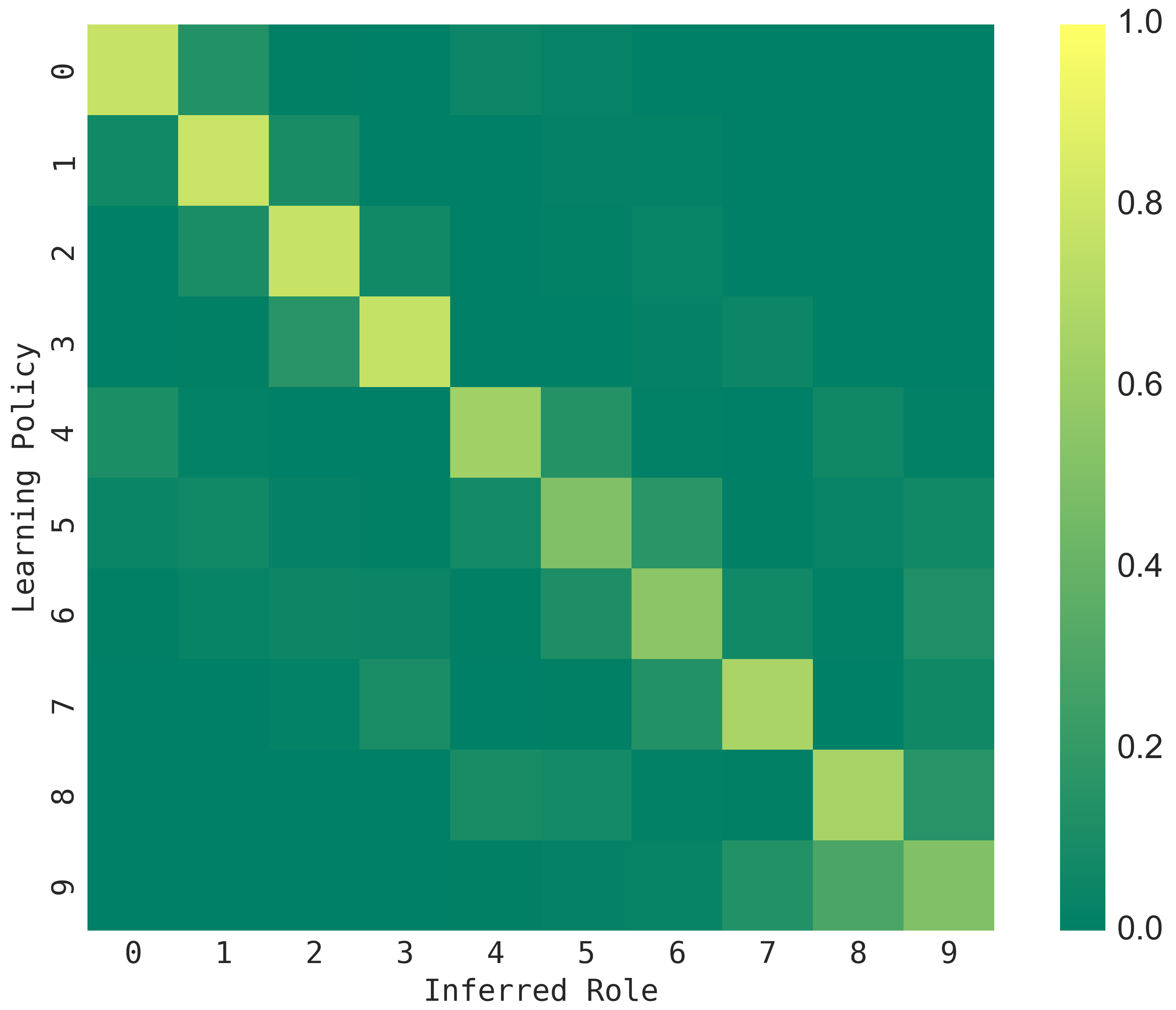}
	\caption{\textit{Role frequency assigned to policy, according to the maximum likelihood estimate of the latent structured model}
		\label{fig:role_frequency}}
\end{figure}

The Gaussian components of latent structure in figure \ref{fig:role_components} give interesting insight about the latent structure of the demonstration data, which correspond to a popular formation arrangement in professional soccer. Unlike the predator-prey domain, however, the players are sometimes expected to switch and swap roles. Figure \ref{fig:role_frequency} displays the tendency that each learning policy $k$ would takes on other roles outside of its dominant mode. 
Policies indexed $0-3$ tend to stay most consistent with the prescribed latent roles. We observe that these also correspond to players with the least variance in their action trajectories. Imitation loss is generally higher for less consistent roles (e.g. policies indexed $8-9$). Intuitively, entropy regularization encourages a decomposition of roles that result in learning policies as decoupled as possible, in order to minimize the imitation loss.